\documentclass[preprint,12pt]{elsarticle}
\usepackage[margin=1in]{geometry}
\usepackage{natbib}
\usepackage{graphicx}
\usepackage{mathrsfs}
\usepackage{color}
\usepackage{amssymb}
\usepackage{amsfonts}
\usepackage{amsmath}
\usepackage{bbm}
\usepackage{nameref}
\usepackage{mathtools}
\usepackage{tikz}
\usepackage{stackengine}
\usepackage{lineno}
\usepackage{setspace}
\usepackage{algorithm}
\usepackage{algpseudocode}
\usepackage{subcaption}

\usepackage{kantlipsum}
\allowdisplaybreaks
\usepackage{mwe}
\usepackage{diagbox}
\usepackage{color, colortbl}
\definecolor{Gray}{gray}{0.9}
\usepackage{hyperref}
\usepackage{placeins}
\usepackage{xcolor} 
\usepackage[utf8]{inputenc}
\usepackage{paralist}
\usepackage{booktabs,caption}
\usepackage[flushleft]{threeparttable}
\usepackage{multirow}
\usepackage{makecell}
\usepackage{csquotes}

\DeclareFontFamily{U}{rcjhbltx}{}
\DeclareFontShape{U}{rcjhbltx}{m}{n}{<->rcjhbltx}{}
\DeclareSymbolFont{hebrewletters}{U}{rcjhbltx}{m}{n}

\usepackage{bbold}

\usepackage{amsthm}

\theoremstyle{definition}

\newtheorem*{definition*}{Definition}

\algtext*{EndFor}
\algtext*{EndIf}
\algtext*{EndFunction}
\algtext*{EndWhile}

\algrenewcommand\algorithmicforall{\textbf{foreach}}
\algrenewcommand\algorithmicindent{.8em}

\begin{document}

\begin{frontmatter}

\title{Fairness-Enhancing Vehicle Rebalancing in the Ride-hailing System}

\author[mitcee]{Xiaotong Guo}
\ead{xtguo@mit.edu}
\author[mitdusp]{Hanyong Xu}
\ead{hanyongx@mit.edu}
\author[mitcee]{Dingyi Zhuang}
\ead{dingyi@mit.edu}
\author[mitdusp]{Yunhan Zheng\corref{cor}}
\ead{yunhan@mit.edu}
\author[mitdusp]{Jinhua Zhao, Ph.D.}
\ead{jinhua@mit.edu}

\address[mitcee]{Department of Civil and Environmental Engineering, Massachusetts Institute of Technology, Cambridge, MA, USA}
\address[mitdusp]{Department of Urban Studies and Planning, Massachusetts Institute of Technology, Cambridge, MA, USA}
\cortext[cor]{Corresponding author}

\begin{abstract}

The rapid growth of the ride-hailing industry has revolutionized urban transportation worldwide. Despite its benefits, equity concerns arise as underserved communities face limited accessibility to affordable ride-hailing services. A key issue in this context is the vehicle rebalancing problem, where idle vehicles are moved to areas with anticipated demand. Without equitable approaches in demand forecasting and rebalancing strategies, these practices can further deepen existing inequities. In the realm of ride-hailing, three main facets of fairness are recognized: algorithmic fairness, fairness to drivers, and fairness to riders. This paper focuses on enhancing both algorithmic and rider fairness through a novel vehicle rebalancing method. We introduce an approach that combines a Socio-Aware Spatial-Temporal Graph Convolutional Network (SA-STGCN) for refined demand prediction and a fairness-integrated Matching-Integrated Vehicle Rebalancing (MIVR) model for subsequent vehicle rebalancing. Our methodology is designed to reduce prediction discrepancies and ensure equitable service provision across diverse regions. The effectiveness of our system is evaluated using simulations based on real-world ride-hailing data. The results suggest that our proposed method enhances both accuracy and fairness in forecasting ride-hailing demand, ultimately resulting in more equitable vehicle rebalancing in subsequent operations. Specifically, the algorithm developed in this study effectively reduces the standard deviation and average customer wait times by 6.48\% and 0.49\%, respectively. This achievement signifies a beneficial outcome for ride-hailing platforms, striking a balance between operational efficiency and fairness. 

\end{abstract}

\end{frontmatter}


\section{Introduction}

Since its introduction in 2009, the ride-hailing industry has witnessed significant global growth. Fueled by technological advancements and the widespread adoption of mobile phones, ride-hailing services offered by Transportation Network Companies (TNCs) like Uber, Lyft, and Didi have revolutionized commuting, creating new economic opportunities. With a market size of approximately 30 billion USD and projected to reach 100 billion USD by 2030~\cite{ride-hailing-market-size}, the industry continues to meet the increasing demand for convenient and flexible transportation options in today's rapidly urbanizing world.

However, alongside the benefits, the ride-hailing industry has also raised significant societal concerns. Research conducted by \citet{diao2021impacts} indicates that the proliferation of TNCs has exacerbated urban mobility challenges, resulting in increased road congestion and decreased usage of public transit. Underserved communities and low-income neighborhoods have been disproportionately affected by the limited accessibility and affordability of ride-hailing services. Additionally, the heavy reliance on algorithms in TNC platforms for tasks such as passenger-driver matching, pricing, and operational optimization poses the risk of perpetuating biases and discrimination if not designed and implemented with fairness in mind.

One of the major operational problems in the ride-hailing system is the vehicle rebalancing problem, where vacant vehicles are redistributed proactively to undersupplied areas to reduce the discrepancy between supply and demand~\cite{spieser2016shared, wallar2018vehicle, Miao_Han_Hendawi_Khalefa_Stankovic_Pappas_2017, wen2017rebalancing, GUO2021161, GUO2022}. Nevertheless, if ride-hailing platforms focus solely on maximizing profits or efficiency without considering equity concerns, their operational approach can trigger a detrimental feedback loop within the system.

\begin{figure}[!h]
\centering
\begin{subfigure}{.48\textwidth}
  \centering
  \includegraphics[width=.95\linewidth]{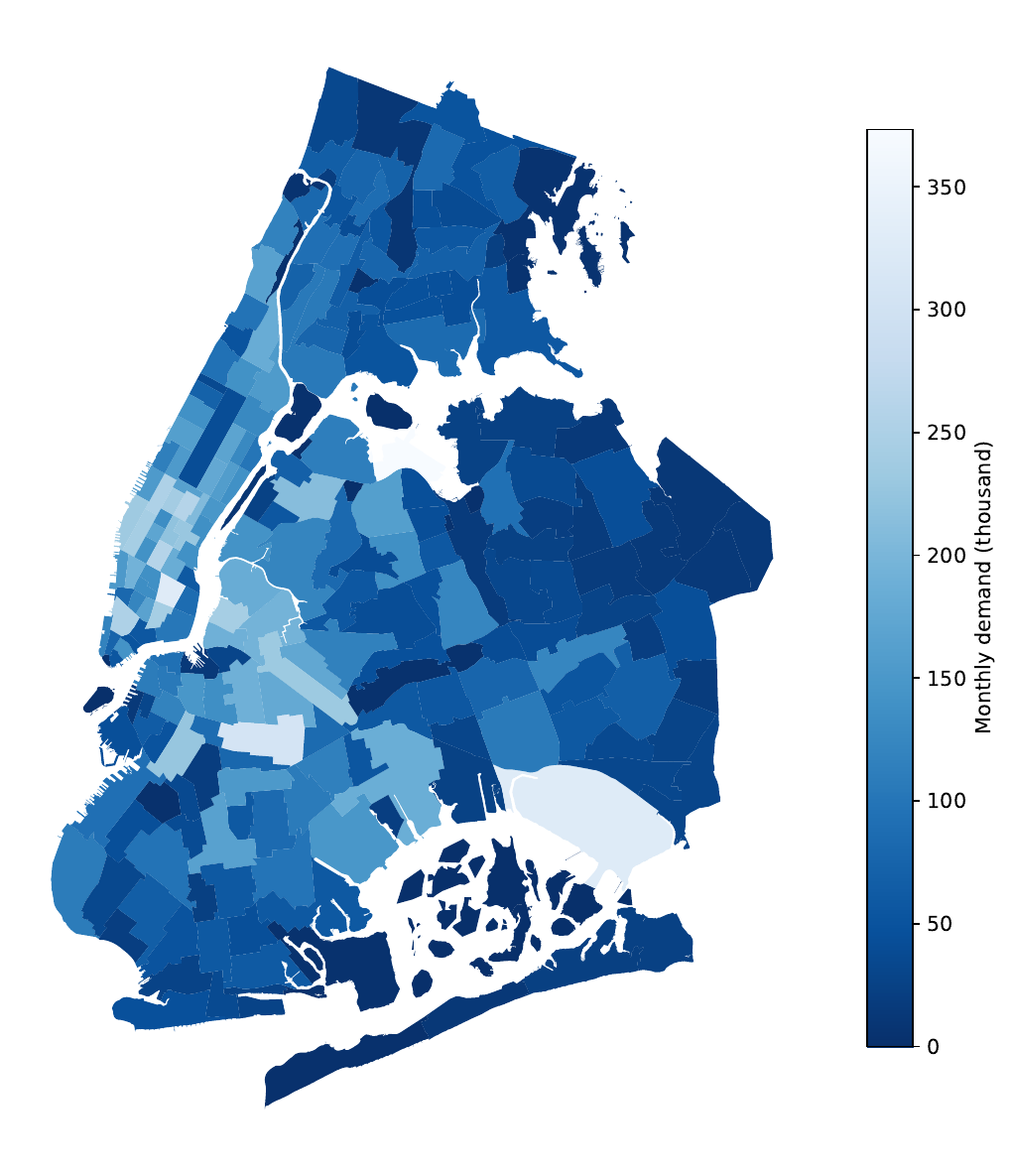}  
  \caption{NYC Ride-Hailing Demand Distribution (by taxi zones)}
  \label{fig:demand}
\end{subfigure}
\begin{subfigure}{.48\textwidth}
  \centering
\includegraphics[width=.95\linewidth]{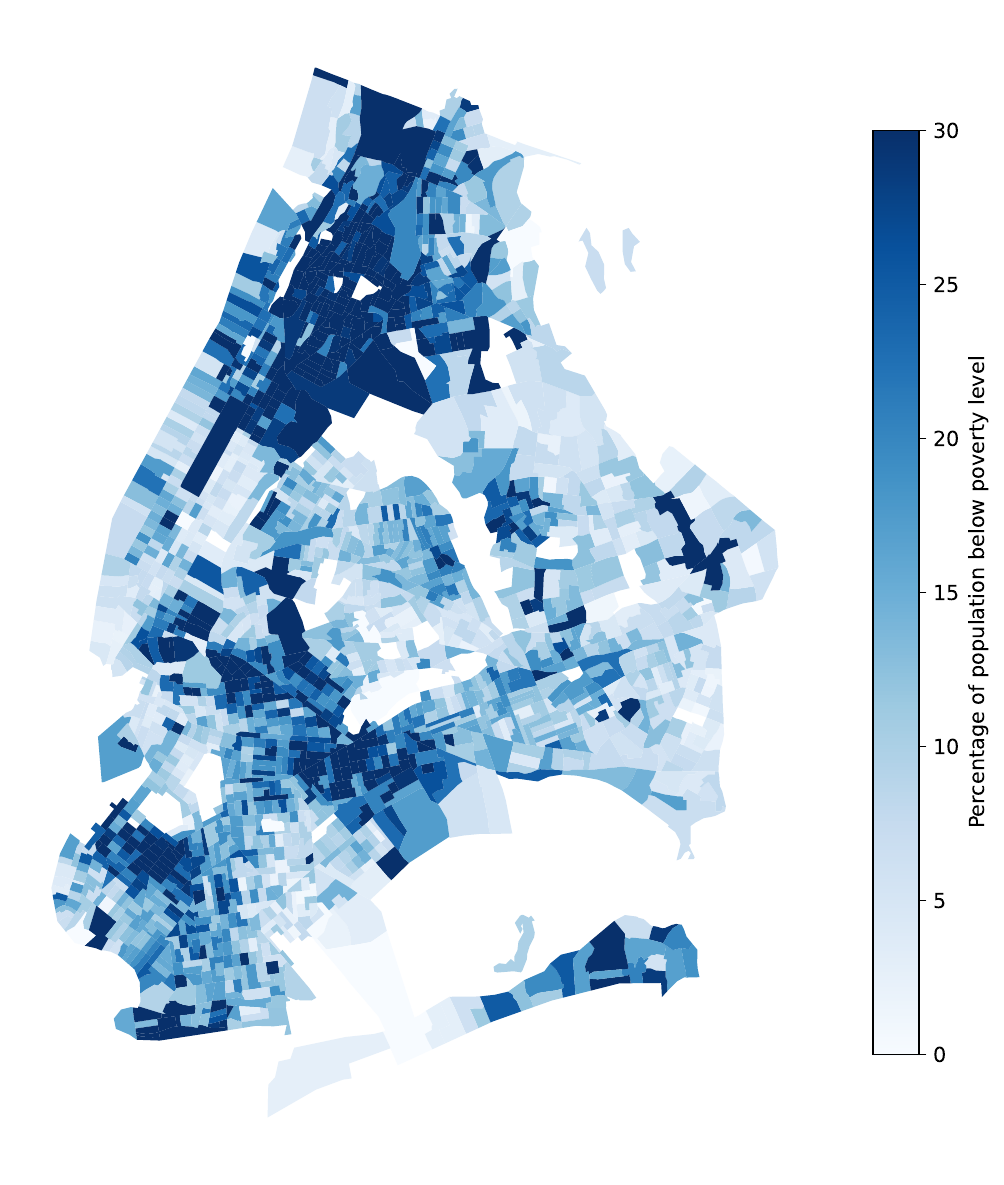}  
  \caption{NYC Poverty Map (by census tracts)}
  \label{fig:poverty}
\end{subfigure}
\caption{Spatial distributions of ride-hailing demand and poverty in New York City (NYC).}
\label{fig:demand_poverty_map}
\end{figure}

Consider the case of New York City (NYC). As illustrated in Figure \ref{fig:demand_poverty_map}, the map displays the spatial distribution of ride-hailing demand and poverty levels in the city. Darker shades represent areas with a higher population living below the poverty line and lower ride-hailing demand. Predictably, regions with a larger impoverished population exhibit reduced ride-hailing demand, such as upper Manhattan, the Bronx, and lower Brooklyn. During the process of redistributing vacant vehicles to meet demand, vehicles tend to be reallocated to high-demand regions, which are more likely to receive future passenger requests. This redistribution, however, diminishes the supply and service levels in low-demand areas, typically inhabited by people facing poverty. Consequently, residents in these low-demand regions are discouraged from using ride-hailing services when the service levels decrease. This negative feedback loop exacerbates the situation, limiting ride-hailing options to affluent individuals and exacerbating serious equity issues.

Moreover, accurate short-term travel demand forecasting is essential for efficient vehicle rebalancing in ride-hailing systems. Data-driven approaches, including traditional time series analysis~\cite{zhang2011seasonal,li2012prediction} and modern machine learning models~\cite{xu2013public,li2015traffic, Ke2017, guo2020residual,li2021multi, wang2023uncertainty}, have been utilized in generating reliable predictions. However, many studies focus solely on prediction accuracy, disregarding the social consequences of travel demand forecasting. Take the vehicle rebalancing problem for instance, under-predicting travel demand in disadvantaged neighborhoods can result in fewer vehicles being dispatched to the area, leading to inadequate services for certain groups. 

Hence, we pinpoint two key fairness concerns in ride-hailing system vehicle rebalancing: firstly, the strategies employed prioritize profit and efficiency at the expense of equity considerations, and secondly, the prediction algorithms essential for rebalancing operations may give rise to algorithmic fairness problems. In this paper, we outline two levels of fairness aimed at improvement:
\begin{enumerate}
    \item \textbf{Fairness in Provisioned Ride-Hailing Services:} Our goal is to reduce disparities in services provided by customers, regardless of the regions from which trips originate.
    \item \textbf{Fairness in Demand Prediction Algorithms:} We aim to minimize prediction errors for each region, irrespective of spatial locations and historical demand levels, involved in vehicle rebalancing operations.
\end{enumerate}

Prior research in the ride-hailing domain has been predominantly focused on enhancing particular facets of fairness: algorithmic fairness, driver fairness, and rider fairness. Notably, algorithmic fairness has been addressed in demand prediction efforts by \citet{Yan_Howe_2020, Yan2021, zheng_fairness-enhancing_2023}. Fairness concerning driver earnings has been explored by \citet{Sühr2019, Bokányi2020, Sun2022}. Additionally, fairness in the service and pricing offered to riders has been discussed in studies by \citet{Nanda2020, Lu2022, Ke2023}. However, it is rare for studies to integrate more than two dimensions of fairness. \citet{Cao2021} has contemplated both the fairness of drivers' income and the fairness to riders concerning varying detour distances. Similarly, \citet{Raman2021} has considered income disparity among drivers as well as the differential treatment of riders based on socio-demographic factors.
Nevertheless, algorithmic fairness typically receives attention as a standalone issue. To the best of authors' knowledge, there have yet to be studies that concurrently address the concerns of algorithmic fairness and rider fairness within the context of ride-hailing systems.

This study introduces a fairness-enhancing vehicle rebalancing framework, taking into account both the algorithmic fairness in demand prediction and the fairness of services provided to riders in the downstream vehicle rebalancing operation, building upon the Matching-Integrated Vehicle Rebalancing (MIVR) algorithm~\cite{GUO2021161}. The framework aims to tackle the two levels of fairness outlined previously. The key contributions of this paper are summarized as follows:

\begin{itemize}
    \item This work represents a novel contribution to the realm of ride-hailing vehicle rebalancing by aiming to advance fairness in the realms of both demand prediction and the provision of services to riders in various regions.
    \item We introduce a Socio-Aware Spatial-Temporal Graph Convolutional Network (SA-STGCN) that builds upon the STGCN model~\cite{yu_spatio-temporal_2018}, enhancing upstream demand prediction. This new framework incorporates a socio-enriched adjacency matrix and a bias-mitigation regularization method to minimize prediction discrepancies across regions.
    \item Building on the MIVR algorithm, we enhance the vehicle rebalancing process with a fairness-focused weighted objective function. This function is informed by the socio-enriched adjacency matrix derived from the SA-STGCN model.
    \item We implement several fairness metrics to evaluate the accuracy of demand prediction algorithms and utilize real-world ride-hailing data to test the fairness of the services provided to riders. Our framework is designed to mitigate fairness issues in both the prediction of demand and the distribution of services to riders.
\end{itemize}

The remainder of this paper is organized as follows. Section \ref{sec:lit_rev} provides a comprehensive review of the existing literature on ride-hailing vehicle rebalancing problems, demand prediction approaches, and fairness issues within the ride-hailing system. In Section \ref{sec:methodology}, we present the fairness-enhanced vehicle rebalancing framework. The results of our numerical experiments are presented in Section \ref{sec:experiments}. Section \ref{sec:policy_discussion} discussed the policy implications from the results. Finally, Section \ref{sec:summary} summarizes the paper and outlines future research directions.

\section{Literature Review} \label{sec:lit_rev}
\subsection{Ride-Hailing System and Vehicle Rebalancing Problem}

The field of ride-hailing systems is extensively studied, as outlined in sources like \citet{Wang_Yang_2019}. This body of research covers a range of topics including the structure of the market~\cite{GUO2023104397, zhang2022economies, wang2023quantifying}, analyses of labor supply~\cite{GUO2023104233}, operations of matching drivers with passengers~\cite{Alonso-Mora_2017}, strategies for vehicle rebalancing~\cite{spieser2016shared, wallar2018vehicle, Miao_Han_Hendawi_Khalefa_Stankovic_Pappas_2017, wen2017rebalancing, GUO2021161, GUO2022}, designs of surge pricing~\cite{Castillo2017}, among other areas. A key operational strategy in these systems is the rebalancing of vacant vehicles. This process is vital in complementing the primary function of connecting customers with available drivers.

A major challenge in ride-hailing systems is the spatial mismatch between where demand arises and where vehicles are available. To address this, there's a need for relocating idle vehicles to areas where future demand is anticipated to exceed the current supply of vehicles. By adopting this proactive rebalancing approach, ride-hailing platforms can significantly reduce the distance traveled by empty vehicles, also known as 'empty miles', and concurrently decrease the waiting times for customers. This strategy is essential for optimizing operational efficiency and enhancing user satisfaction in ride-hailing services.

The vehicle rebalancing problem is initially studied by \citet{Godfrey_Powell_2002_A, Godfrey_Powell_2002_B} under the context of dynamic fleet management. Over the past decade, with the rapid growth of Mobility-on-Demand (MoD) and ride-hailing systems, more attention has been devoted to solving this challenge~\cite{wen2017rebalancing, Jiao_Tang_Qin_Li_Zhang_Zhu_Ye_2021, Braverman_Dai_Liu_Ying_2019, Miao_Han_Hendawi_Khalefa_Stankovic_Pappas_2017}. \citet{wen2017rebalancing} used reinforcement learning to tackle vehicle rebalancing in a shared MoD system, achieving a 14\% fleet size reduction in a London simulation. \citet{Braverman_Dai_Liu_Ying_2019} designed a fluid-based optimization model for ride-hailing vehicle management, resulting in improved passenger service compared to benchmark models. \citet{Miao_Han_Hendawi_Khalefa_Stankovic_Pappas_2017} introduced a data-driven vehicle rebalancing model, minimizing the worst-case rebalancing cost using real-world NYC taxi data, achieving an average 30\% reduction in idle driving distance. \citet{GUO2021161} proposed the Matching-Integrated Vehicle Rebalancing (MIVR) model for solving the vehicle rebalancing problem considering future iterations and incorporated the demand uncertainty with the Robust Optimization (RO) techniques. \citet{GUO2022} expanded on this concept by exploring multiple data-driven strategies to handle demand uncertainty within the framework of the MIVR model. 

Meanwhile, various studies have been focused on the control of Autonomous MoD system~\cite{Zardini_Lanzetti_Pavone_Frazzoli_2021, Pavone_Smith_Frazzoli_Rus_2012, Zhang_Pavone_2014, Iglesias_Rossi_Wang_Hallac_Leskovec_Pavone_2017, Tsao_Milojevic_Ruch_Salazar_Frazzoli_Pavone_2019}, where vehicles are dispatched in the system by the proposed control strategy. \citet{Pavone_Smith_Frazzoli_Rus_2012} used a fluid-based model and linear program for generating optimal rebalancing policy. \citet{Zhang_Pavone_2014} proposed a queueing-based algorithm for AMoD rebalancing. \citet{Iglesias_Rossi_Wang_Hallac_Leskovec_Pavone_2017} utilized LSTM neural networks in a Model Predictive Contro (MPC) algorithm for rebalancing with short-term demand forecasts. \citet{Tsao_Milojevic_Ruch_Salazar_Frazzoli_Pavone_2019}  further introduced an MPC algorithm in the shared AMoD setting.

Despite the extensive research on vehicle rebalancing, none of the existing studies have explicitly tackled equity concerns in their proposed algorithms. This paper aims to fill this gap by presenting a fairness-enhanced vehicle rebalancing framework building upon the MIVR model~\cite{GUO2021161} that systematically addresses issues related to the perceived services experienced by customers.

\subsection{Demand Prediction in the Transportation System}
Predicting the accurate travel demand of a given transportation system is crucial for efficient traffic operations and regulations. A great volume of research has studied diverse methods for travel demand forecasts. Traditional common methods include the Historical Averages, Moving Averages, autoregressive integrated moving average (ARIMA) and its variants, and some basic machine learning models such as support vector machines (SVM) \cite{zhang2011seasonal, Jeong_Byon_Castro-Neto_Easa_2013, lippi_short-term_2013, vlahogianni_short-term_2014}. However, these models focused more on temporal links but failed to account for spatial and relational information in the transportation network \cite{lippi_short-term_2013}.

In recent years, with the rise of large machine learning models and the booming computing power, there has been a shift from using traditional statistical time series analysis to deep learning sequential networks \cite{vlahogianni_short-term_2014}. Many studies have leveraged the convolutional neural networks (CNN) such as ResNet to capture spatial features \cite{Zhang_Zheng_Qi_Li_Yi_2016, Zhang_Zheng_Qi_2017} and used the recurrent neural networks (RNN) such as the long-short term memory (LSTM) and Attention mechanism to learn the time series function \cite{Yao_Tang_Wei_Zheng_Li_2019, Ke2017, Yao_Wu_Ke_Tang_Jia_Lu_Gong_Ye_Li_2018}. 

On the other hand, graph neural networks (GNN) gained popularity because of their ability to capture spatial dependency and the non-Euclidean structure of the street network \cite{jiang_graph_2022}. There are many types of graph convolutions in GNNs, including the spectral-based graph convolutional networks (GCN) \cite{kipf_semi-supervised_2017, defferrard_convolutional_2017} and spatial-based convolutional GNNs \cite{Atwood_Towsley_2016, Gilmer_Schoenholz_Riley_Vinyals_Dahl_2017, Hamilton_Ying_Leskovec_2017, Veličković_Cucurull_Casanova_Romero_Liò_Bengio_2018}. Many architectures are proposed to solve traffic forecasting problems. For instance, \citet{Li_Yu_Shahabi_Liu_2018} proposed the Diffusion Convolutional Recurrent Neural Network (DCRNN), leveraging bidirectional graph random walk and the recurrent neural network to capture both the spatial and temporal dependencies. \citet{yu_spatio-temporal_2018} proposed the Spatial-Temporal Graph Convolutional Network (STGCN), using graph and temporal convolution layers to build up the basic block of the architecture. \citet{Wu_Pan_Long_Jiang_Zhang_2019} developed the Graph WaveNet, which utilized a self-adaptive adjacency matrix to capture hidden relations in the graph and leveraged dilated causal convolution to work with long-range sequences. Our research adopted the STGCN model as the main structure for its state-of-art performance and ease of implementation. 

\subsection{Fairness in the Transportation System}
\subsubsection{Fairness Definition}

There is a wide range of debates regarding the definition of fairness in political philosophy, computer science, and transportation. Major theories of fairness in political philosophy can be separated into four categories, which are 1) ensure equal share or proportional share for each individual, 2) ensure market equilibrium, 3) maximize total welfare, and 4) ensure subgroup welfare \cite{lewis_exploring_2021, binns_fairness_2018}. In computer science, there are also various evaluation metrics to measure algorithmic fairness, the first set is Disparate Impact Analysis versus Disparate Treatment Analysis, where the former aims to achieve fair impact or results for the unit of comparison while the latter aims to achieve fair treatment \cite{mehrabi_survey_2022, pessach_review_2022}. Another set of evaluation notions based on the unit of comparison includes group-based and individual-based fairness, where the former focuses on the same outcome or treatment of different groups, while the latter concentrates on the same outcome or treatment of individuals \cite{caton_fairness_2020, mehrabi_survey_2022}. In transportation, horizontal equity and vertical equity are often alluded to. Horizontal equity refers to the goal of similar people receiving similar treatment, while vertical equity refers to the goal of disadvantaged people being taken care of \cite{litman_evaluating_nodate, Yan_Howe_2020}. This fairness notion aligns with the many existing concepts of fairness and equity from the fields of political philosophy and computer science.

\subsubsection{Fairness Research in the Transportation System}

Research in fairness and equity in the transportation system has been emerging in recent years, especially in public transit planning, where the fair sharing of public resources is important. Previously, a lot of the focus was put on the equity analysis of the existing or forthcoming systems. \citet{bills_looking_2017} measured and compared consumer surplus distributions for different population segments across planning scenarios in the Bay area. \citet{cascetta_economic_2020} estimated the horizontal equity of the travel time accessibility by calculating the Gini index for Italian high-speed railways. On the other hand, \citet{zheng_equality_2021} demonstrated the unfairness in the Deep Neural Network and the Discrete Choice Model, unlike preceding research, they also provided a disparity mitigation solution through an absolute correlation regularization method.

In the realm of ride-hailing, researchers have been focused on improving algorithmic fairness, driver fairness, and rider fairness. 
\citet{Yan_Howe_2020} addressed the challenge of socio-economic inequity perpetuated by new mobility services like car-sharing, bike-sharing, and ride-hailing.
\citet{Yan2021} discussed a novel unsupervised learning architecture, named EquiTensors, for integrating heterogeneous spatio-temporal urban data to counteract bias and produce fair and reusable representations.
Both \citet{zheng_fairness-enhancing_2023} and \citet{zhang_enhancing_2023} have looked at the demand prediction fairness in ride-hailing, however, they only leveraged group fairness as a fairness definition and focused on the prediction result without considering the downstream impact. 

In exploring driver-side fairness in ride-hailing systems, various studies have made significant contributions. \citet{Bokányi2020} employed a city simulation to highlight critical issues of fairness and wage inequality. \citet{Sühr2019} analyzed drivers' income fairness over time, suggesting a framework that enhances the utility for both customers and drivers in a dual-sided market. \citet{Sun2022} tackled the dual challenge of efficiency and fairness, introducing a multi-agent reinforcement learning framework that assists drivers in making equitable income decisions through order selection and repositioning. \citet{Raman2021} offered two strategies within a Markov decision process to mitigate inequality in ride-pooling services, aiming to balance profitability and fairness for both passengers and drivers.

As for rider-side fairness in transportation systems, \citet{qian_scram_2015} developed a novel route recommendation system that considers shared roads while maintaining cost-effectiveness for each customer. \citet{Cao2021} aimed at enhancing both efficiency and fairness in ride-sharing by optimizing routes for ride-hailing services. \citet{Ke2023} introduced the concept of Fleet-Optimal Behavior with Service Constraint (FOSC), focusing on striking a balance between reducing total fleet costs and ensuring fair travel times for riders. \citet{Nanda2020} proposed a driver-customer matching algorithm that addresses the challenge of maximizing profit while maintaining fairness in rider matching rates. \citet{Lu2022} suggested a decentralized smart price auditing system using block-chain technology and smart contracts to ensure price fairness and prevent discrimination among riders in ride-hailing services.

In this paper, we present a new vehicle rebalancing framework aimed at simultaneously enhancing algorithmic and rider fairness, aspects not jointly considered in existing research. This innovative approach ensures equitable and efficient operations, setting a new standard for fairness in vehicle rebalancing.

\section{Methodology} \label{sec:methodology}

In this section, we will introduce the fairness-enhancing vehicle rebalancing framework, consisting of a Socio-Aware STGCN (SA-STGCN) demand prediction component and a fairness-enhancing MIVR component for downstream vehicle rebalancing operations.

\subsection{Fairness-Enhancing Demand Prediction Framework}

The original STGCN (Spatio-Temporal Graph Convolutional Network) framework, while effective in numerical prediction accuracy, primarily concentrated on leveraging spatial locality information and temporal autocorrelations. This approach, though beneficial in certain contexts, inadvertently leads to a regional imbalance in the model outputs. Such imbalances can propagate unfairness in downstream applications, particularly in tasks that are sensitive to regional disparities. Recognizing this limitation, our research proposes the innovative SA-STGCN (Social Aware STGCN) framework, complemented by a fairness-enhancing methodology specifically tailored for vehicle rebalancing optimization in ride-hailing operations. The conceptual architecture of this approach is detailed in Figure \ref{fig:diagram}:

\begin{figure}[htbp]
\centering
\includegraphics[width=0.9\textwidth]{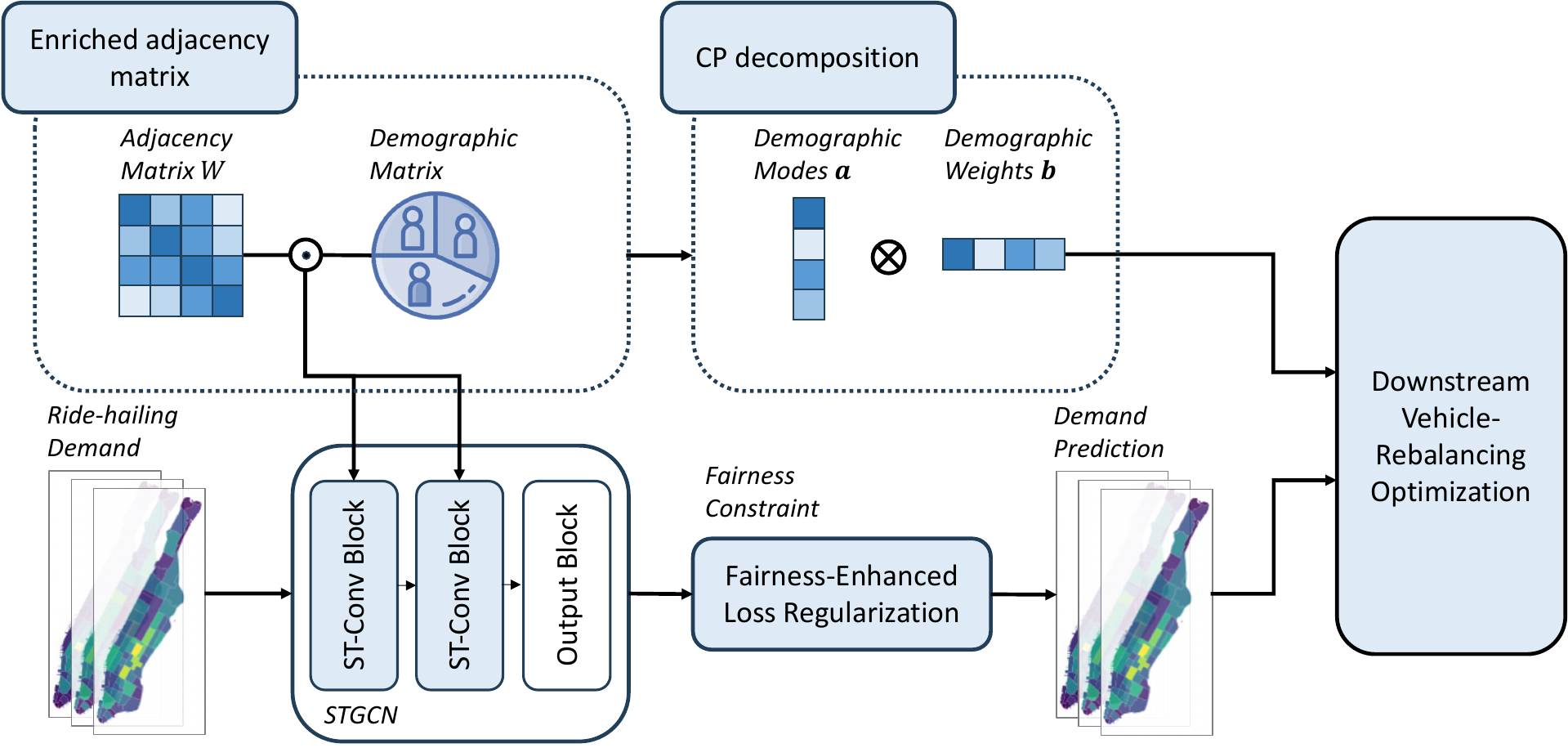}
\caption{Detailed illustrations of the SA-STGCN framework and its integration with the vehicle rebalancing optimization task.}
\label{fig:diagram}
\end{figure}

As depicted in Figure \ref{fig:diagram}, our modifications to the STGCN framework are multifaceted and designed to address the fairness issue at its core. These modifications include:
\begin{enumerate}
    \item Implementing a fairness-enhanced loss regularization approach to penalize demand overestimations and reduce output disparities.
    \item Integrating a demographic matrix (e.g., representing the Black and low-income population) with the original adjacency matrix to infuse social context into the data.
    \item Employing matrix decomposition on the augmented adjacency matrix to derive demographic weights, which serve as adjustment factors in the downstream rebalancing task.
\end{enumerate}

In the following sections, we delve deeper into each of these components, discussing their implementation and the specific methodologies employed. We aim to provide a comprehensive understanding of how each aspect contributes to the overall effectiveness and fairness of the SA-STGCN framework.

\subsubsection{Socio-Aware Spatial-Temporal Graph Convolutional Network}
This module starts by formulating the demand prediction problem. Given the ride-hailing system, we can regard it as a directed weighted graph $G = (V, E, W)$, where $|V| = n$ is the set graph vertices representing sub-regions (zones), the $E$ represents the set of road networks between each pair of vertices as edges, and $W$ represents the weights of each edge in the form of weighted adjacency matrix, calculated by:
\begin{linenomath}
\begin{equation}
W_{ij} = \exp{( - \frac { \Bar{d}_{ij} ^2}{\sigma_{d}^2})},
\end{equation}
\end{linenomath}
where $W_{ij}$ is the edge weight between graph vertices $v_i$ and $v_j$, $\Bar{d}_{ij}$ is the Euclidean distance between the centroids of vertices $v_i$ and $v_j$, $\sigma_{d}$ is the standard deviation of the set of distances begins at each vertex $v_i$ \cite{defferrard_convolutional_2017}. 

To capture the spatial demographic features that exist in the network, this study establishes a socio-demographically enriched adjacency matrix by incorporating an additional demographic matrix. In this study, we leverage census data and focus on the ratio of the minority race population and the population in poverty as the variables of interest. The first step is to construct a demographic matrix. Consider the demographic variables in preceding $N$ years in each graph vertice $v_i$ as a vector $z_i$ where $|z_i| = 2N$ since we only focused on two variables here. Then we can construct the relationship between the demographic features of each pair of vertices $z_i$ and $z_j$ with a correlation matrix. 

\begin{linenomath}
\begin{equation}
    Corr_{ij} = \frac{\sum^{2N}_{m=1}(z^m_i - \Bar{z_i})(z^m_j - \Bar{z_j})}{\sqrt{\sum^{2N}_{m=1}(z^m_i - \Bar{z_i})^2}\sqrt{\sum^{2N}_{m=1}(z_j - \Bar{z_j})^2}},
    \label{eq:demo_correlation}
\end{equation}
\end{linenomath}

where $z^m_i$ and $z^m_j$ are the individual elements in $z_i$ and $z_j$. When $z_i$ = $z_j$ = 0, the corresponding value in the matrix will be filled as zero.

We derived a new socio-demographically adjacency matrix that incorporates this demographic matrix into the original adjacency matrix, specifically, we use the Hadamard product of the original adjacency matrix and the demographic correlation matrix as the new adjacency matrix to feed into the prediction model described below. To introduce sparsity, the element value of the new matrix is only kept when it is greater than or equal to 0.1, otherwise, it is converted to 0.

\begin{linenomath}
\label{eq:enriched_adjacency}
\begin{equation} 
W_{ij}^* = 
    \begin{cases}
        W_{ij} \circ Corr_{ij}, & \text{if } W_{ij} \circ Corr_{ij} \geq 0.1, \\
        0, & \text{otherwise}.
    \end{cases}
    \label{eq:enriched_adj}
\end{equation}
\end{linenomath}

In terms of the representation of the ride-hailing demand, let $r_i^k\in \mathbb{N}$ be the demand at each time period $k$ that originates at vertex $v_i$ (or sub-region $i$). Given the demand observations from previous $M$ time periods at each vertex $[r^{k-M}, ..., r^{k-1}]$, we want to forecast the upcoming demand $\hat{r}^{k+1}$.

The study refers to the Spatial-Temporal Graph Convolutional Network (STGCN) proposed by \citep{yu_spatio-temporal_2018} as the main model structure. STGCN has been widely used in traffic forecasting for its excellent ability to capture temporal and spatial features compared to traditional CNN and time-series models because it utilizes both graph and gated temporal convolutions. The model we adopted consists of two Spatial Temporal Convolutional (ST-Conv) blocks followed by a fully connected layer. Each ST-Conv block comprises two temporal gated convolution layers and a spatial graph convolution layer in between. To reduce the computing cost of graph convolution, we adopted the Chebyshev Polynomials Approximation and the 1st-Order Approximation strategies \cite{defferrard_convolutional_2017, kipf_semi-supervised_2017, yu_spatio-temporal_2018}.

\subsubsection{Fairness Enhancement}

This research takes the individual fairness definition mentioned previously as the objective. In this study, we addressed the fairness issue with regularization inspired by the work of \citep{zheng_fairness-enhancing_2023} and \citep{Beutel_Chen_Doshi_Qian_Woodruff_Luu_Kreitmann_Bischof_Chi_2019} and applied three versions of fairness cost terms in the cost function. The set of demand predictions $\hat{R}$ and all the trainable parameters in the model $W_{\theta}$ gives the cost function. In general, the cost function consists of two parts, each optimizes the accuracy and the fairness of the model performance:
\begin{linenomath}
\begin{equation}
    J_{total} = J_{accuracy} + param * J_{fariness},
\end{equation}
\end{linenomath}
where $param$ is a hyperparameter that adjusts the relative weight of the fairness optimization term. The research uses $\lambda$ and $\gamma$ as the weight parameter for the case of limiting error distribution and penalizing overestimation, respectively. 

\textbf{Limit error distribution.} To achieve fairness for the whole population and regulate the error distribution variance, we add the variance of the symmetric absolute percentage error (SAPE) to the cost function. First, $SAPE$ at each vertex in a given time period is defined as:
\begin{linenomath}
    \begin{equation}
    SAPE_i^k = 
        \begin{cases}
        \frac{|r_{i}^k - \hat{r}_{i}^k|}{|r_{i}^k| + |\hat{r}_{i}^k|}, & |r_{i}^k| + |\hat{r}_{i}^k| \neq 0,\\
        0, & \text{otherwise},
        \end{cases}
    \end{equation}
\end{linenomath}
where $r_{i}^k$ and $\hat{r}_{i}^k$ are the original demand and the predicted demand in vertex $v_{i}$ at time $k$. Measuring the percentage error in this way is useful in the case of ride-hailing as the prediction and original demand are sometimes equal or close to zero, making the traditional percentage error term fraction undefined. In the case where both the prediction and original demand are zero, we set the error term as zero. Then, we can calculate the variance of the $SAPE$ for all vertices in a given time period to the cost function:
\begin{linenomath}
    \begin{equation}
        J(\hat{R}, W_{\theta}) = 
        \sum_{k=1}^{\kappa} \sum_{i=1}^{n} (r_{i}^k - \hat{r}_{i}^k)^2 + 
        \lambda \sum_{k=1}^{\kappa} \frac{\sum_{i=1}^n (SAPE_{i}^k - \Bar{SAPE^k})^2}{n-1},
    \end{equation}
\end{linenomath}
where $\kappa$ is the number of time periods in each training iteration.
    
\textbf{Penalize overestimation.} As \citet{GUO2022} pointed out in their study of ride-hailing vehicle rebalancing operations, when the model prediction is not accurate enough, underestimation is preferred compared to overestimation because this avoids unnecessary vehicle relocation, resulting in a better rebalancing result. Moreover, the resultant smaller overestimation will help to reduce the total variance of the prediction error, thus providing a fairer outcome. To reduce overestimation, we add a penalizing term to the cost function:
\begin{linenomath}
    \begin{equation}
        J(\hat{R}, W_{\theta}) = 
            \sum_{k=1}^{\kappa} \sum_{i=1}^{n} (r_{i}^k - \hat{r}_{i}^k)^2 + 
        \gamma \sum_{t=1}^{\kappa} \sum_{i=1}^{n} max(0, \hat{r}_{i}^k - r_{i}^k).
    \end{equation}
\end{linenomath}    
In this way, whenever the prediction overestimates the demand, that is, when $\hat{r}_{i}^k > r_{i}^k$, the regularization term will penalize it to be closer to zero.
    
\textbf{Limit error distribution and penalize overestimation.} Lastly, we combine the previous two regularizations together to achieve the effect of both limiting error distribution and penalizing overestimation:
\begin{linenomath}
\small
\label{eq:demand_obj}
    \begin{equation}
        J(\hat{R}, W_{\theta}) = 
        \sum_{t=1}^{\kappa} \sum_{i=1}^{n} (r_{i}^k - \hat{r}_{i}^k)^2 + 
        \lambda \sum_{k=1}^{\kappa} \frac{\sum_{i=1}^n (SAPE_{i}^k - \Bar{SAPE^k})^2}{n-1} +
        \gamma \sum_{k=1}^{\kappa} \sum_{i=1}^{n} max(0, \hat{r}_{i}^k - r_{i}^k)
    \end{equation}
\end{linenomath}  

\subsubsection{Weighted Adjacency Matrix Decomposition}

The first two components of the SA-STGCN framework focus on addressing fairness concerns in demand prediction. This section introduces a matrix decomposition method to create fairness weights for sub-regions, indicating the importance of repositioning idle vehicles to each sub-region. This vector is then applied in the subsequent vehicle rebalancing task, aiming to enhance fairness in the delivery of ride-hailing services.

In order to obtain the fairness weights of different regions to guide the downstream rebalancing task, we decomposed our self-designed adjacency matrix to obtain a set of fairness weights that consider both the spatial dependencies as well as the disparity of socio-demographic features in the city. Therefore, two consecutive steps are conducted through the process: (1) Enrich the weighted adjacency matrix $W$ with sociodemographic information, which was explained by Equation~\ref{eq:demo_correlation} and \ref{eq:enriched_adj}; (2) Decompose the adjacency matrix to obtain the spectrum and use it as the output weights, which will be explained in this section.

The decomposition of the adjacency matrix is usually applied in the graph spectrum analysis because it provides a powerful means to reveal and quantify the latent structural properties of graphs, such as node centrality, community structure, and connectivity patterns \citep{cvetkovic2009introduction,grone1990laplacian}. In our context, we use Canonical polyadic (CP) decomposition in our experiment. The CP decomposition, also known as PARAFAC or tensor factorization, is a multilinear algebraic framework that generalizes the matrix singular value decomposition to higher-order tensors \citep{hitchcock1927expression,harshman1970foundations,wang2023low}. For our adjacency matrix $W^*$, viewed as a two-dimensional tensor, CP decomposition factorizes $W^*$ into a sum of component rank-one tensors, providing insights into multi-way interactions. Mathematically, if $W^*$ is a tensor of order two, the CP decomposition is represented as:

\begin{equation}
W^* \approx \sum_{r=1}^{R} \lambda_r \mathbf{a}_r \otimes \mathbf{b}_r,
\end{equation}
where $\otimes$ denotes the outer product, $R$ is the rank of the decomposition, $\lambda_r$ are the weights indicating the importance of each component, and $\mathbf{a}_r$ and $\mathbf{b}_r$ are the corresponding factor vectors representing the demographic modes and weights of the decomposed $W^*$. This decomposition facilitates the distillation of complex network data into a form that accentuates inherent spatial and sociodemographic relationships, allowing for compressing $W^*$ to capture nuanced regional disparities in a comprehensible manner. In this study, we define \( R = 1 \) to produce a pair of vectors that encapsulate the maximum amount of information from the adjacency matrix \( W^* \).

The fairness weights for the downstream optimization problem will be derived from the vector \( \boldsymbol{b} \) by following these procedures: i) Normalize \( \boldsymbol{b} \) using a min-max scaling method, ii) decrease the magnitude of each element by multiplying with 0.1, and iii) obtain the final fairness weights for each sub-region by subtracting each scaled value from 1. Specifically, for each sub-region \( i \), the normalized value \( \Bar{b_i} \) is computed as 
\[ \Bar{b_i} = \frac{b_i - \min(\boldsymbol{b})}{\max(\boldsymbol{b}) - \min(\boldsymbol{b})}. \]
Subsequently, the fairness weight for each sub-region \( i \) is determined as 
\[ \omega_i = 1 - 0.1 \times \Bar{b_i}. \]

Our enriched adjacency equation (Equation \ref{eq:enriched_adjacency}) indicates that sub-regions nearer to other sub-regions, with lesser poverty rates and smaller minority race populations, are likely to have higher \( b_i \) values. In contrast, city outskirts, typically marked by higher poverty and more significant minority race populations, are distinct from urban centers. The aim of the fairness weights is to prioritize these underserved peripheral areas, which often face greater socio-economic challenges. These weights are calculated following the earlier mentioned steps, ensuring that sub-regions with elevated \( b_i \) values correspondingly have lower \( \omega_i \) values, thus aligning with our objective of equitable distribution.

\subsection{Fairness-Enhancing Vehicle Rebalancing Component}

Tackling fairness concerns in the demand prediction module doesn't directly mitigate fairness issues related to the services received by customers, who are the primary indicators when evaluating ride-hailing operations. Therefore, this provides a fairness-enhancing Matching-Integrated Vehicle Rebalancing (MIVR) model, building upon the MIVR model proposed by Guo et al.~\cite{GUO2021161}. 

The operational period is divided into $\Omega$ identical time intervals, each denoted by an index $k = 1, 2, ..., \Omega$, and lasting $\Delta$ time units. 
Furthermore, the study area is partitioned into $n$ sub-regions (zones), with each sub-region $i$ exhibiting an estimated demand $r_i^k \geq 0$ at time $k$. To formulate the model, we introduce two sets: i) the set of sub-regions denoted as $N = {1, 2, ..., n}$, and ii) the set of time intervals represented by $K = {1, 2, ..., \kappa}$.

The MIVR model is solved in a rolling-horizon manner: when solving the MIVR model at the start of time interval $k$, it considers the demand during time interval $k$ as well as the demand for $\kappa$ future time intervals; however, only the vehicle rebalancing decisions for the current time interval $k$ are put into action; following this, vehicle locations are observed and updated as inputs to the MIVR model for the subsequent time interval.

The MIVR model consists of two components: matching and rebalancing. The matching component uses decision variables $y_{ij}^k \in \mathbb{R}^+$ to represent the number of customers matched between sub-regions $i$ and $j$ at time $k$. The rebalancing component uses decision variables $x_{ij}^k \in \mathbb{R}^+$ to denote the number of idle vehicles rebalanced from sub-region $i$ to sub-region $j$ at time $k$. Travel distance $d_{ij}^k$ between sub-regions $i$ and $j$ at time $k$ is approximated by the distance between their centroids. 

The fairness considerations are incorporated by introducing the fairness weights $\boldsymbol{\omega} \in (\mathbb{R}^+)^n$, consisting of a weight parameter $\omega_i$ for each region $i$. The fairness weights $\boldsymbol{\omega}$ are decomposed from the enriched adjacency matrix from the SA-STGCN framework. The weighted objective function of the MIVR model can then be formulated as

\begin{align}
\small
\label{eq:MIVR}
    \min_{\boldsymbol{x}, \boldsymbol{y}} \quad & c(\boldsymbol{x}, \boldsymbol{y};\boldsymbol{r}) =  \sum_{k=1}^\kappa \sum_{i=1}^n \sum_{j=1}^n x_{ij}^k d_{ij}^k + \alpha \cdot \sum_{k=1}^\kappa \sum_{i=1}^n \sum_{j=1}^n \omega_i y_{ij}^k d_{ji}^k + \beta \cdot \sum_{k=1}^\kappa \sum_{i=1}^n \omega_i \left(r_i^k - \sum_{j=1}^n y_{ij}^k \right),
\end{align}
where $\boldsymbol{r}$ stands for the vector of estimated demand, $\alpha$ and $\beta$ are weight parameters indicating weights for matching distance and penalty for unsatisfied demand, respectively. The objective function $c(\boldsymbol{x}, \boldsymbol{y};\boldsymbol{r})$ defines a weighted generalized cost for the vehicle rebalancing problem with the consideration of the matching component. 
The fairness weights $\boldsymbol{\omega}$ play a crucial role in rebalancing decisions by imposing extra costs and penalties on specific regions, thereby guiding vehicles towards these areas. The constraints used in this optimization problem can be found in \citet{GUO2021161}. 

With a given demand prediction $\boldsymbol{r}$, problem \ref{eq:MIVR} provides a vehicle rebalancing strategy for the upcoming decision time interval. The demand prediction $\boldsymbol{r}$ is provided by the upstream SA-STGCN framework, which balances the prediction accuracy and fairness metrics.

\section{Experimental Results} \label{sec:experiments}
\subsection{Data}

This research utilizes For-Hire Vehicle trip records obtained from the \citet{NYCTLC}. The experiments are conducted using ``taxi zones'', which are well-defined regions within the high-volume ride-hailing trip dataset, comprising a total of 63 zones on Manhattan Island.

\begin{figure}[!h]
\centering
\begin{subfigure}{.35\textwidth}
  \centering
  \includegraphics[width=.95\linewidth]{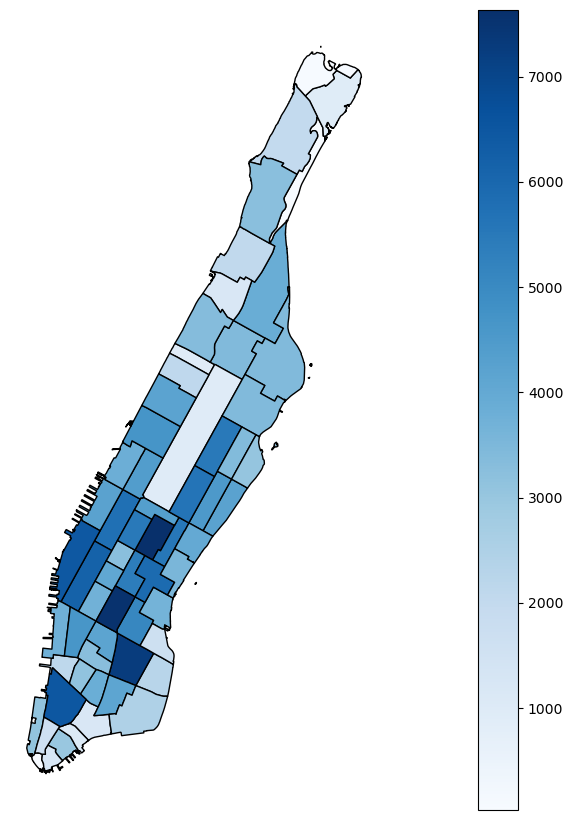}  
  \caption{Mean}
  \label{fig:mean}
\end{subfigure}
\begin{subfigure}{.35\textwidth}
  \centering
\includegraphics[width=.95\linewidth]{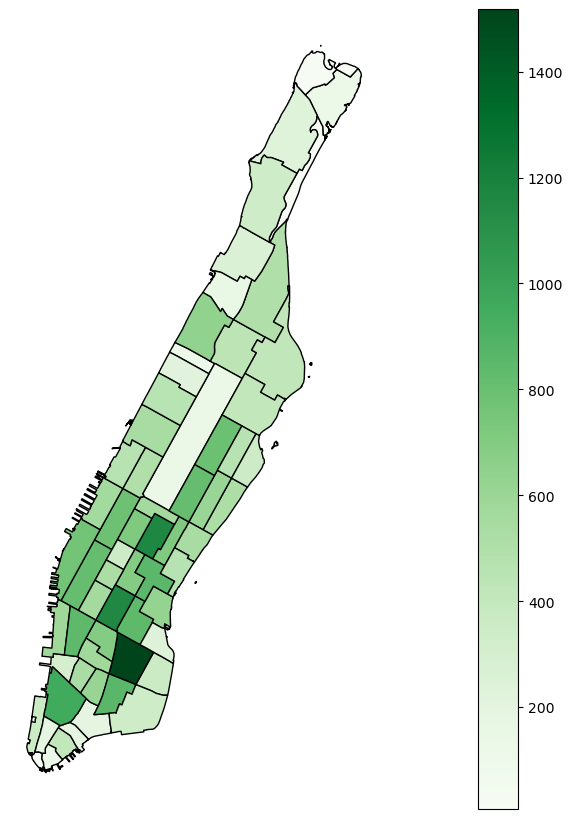}  
  \caption{Standard deviation}
  \label{fig:std}
\end{subfigure}
\caption{Daily demand by zone (trips) in Manhattan.}
\label{fig:peak_demand}
\end{figure}

To best emulate typical travel behaviors, five months of demand data from February to June 2019 are utilized, encompassing a total of 47,009,841 trips within the study period. The demand data is aggregated into 5-minute time intervals ($\Delta = 300$ seconds) for analysis and evaluation. 
Figure \ref{fig:peak_demand} illustrates the average and standard deviation of daily regional demand in Manhattan, revealing significant demand volatility in the lower Manhattan area.

In the demand prediction module, these taxi zones are employed as vertices in the graph $G$. The dataset is divided into three subsets: the training set, validation set, and test set, which account for 70\%, 15\%, and 15\% of the overall dataset, respectively. To predict the demand $r^{k}$ for a given taxi zone at time $k$, the module utilizes the previous 12 observations of demand $[r^{k-12}, r^{k-11}, ..., r^{k-1}]$.

To assess the performance of both the demand prediction and vehicle rebalancing modules, real-world demand data from June 26, 2019, is used as the evaluation set. This one-day demand data is excluded from the training stage to ensure a fair assessment of the modules' capabilities.

\begin{figure}[!h]
\centering
\begin{subfigure}{.35\textwidth}
  \centering
  \includegraphics[width=.95\linewidth]{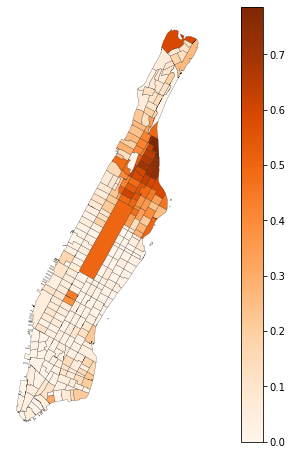}  
  \caption{Black Population Ratio}
  \label{fig:demo_black}
\end{subfigure}
\begin{subfigure}{.35\textwidth}
  \centering
\includegraphics[width=.95\linewidth]{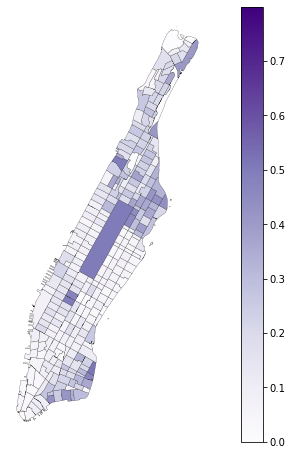}  
  \caption{Poverty Population Ratio}
  \label{fig:demo_poverty}
\end{subfigure}
\caption{Demographic Variables Distribution in Manhattan in 2019.}
\label{fig:demographic}
\end{figure}

To incorporate the socio-demographic information into both the upstream prediction and downstream operation, we used the American Community Survey (ACS) census data. Specifically, we focused on two variables, the racial and poverty compositions of a given region to represent the population vulnerable to insufficient ride-hailing service provision. The poverty level is pre-determined by the ACS dataset. We also selected five years of data from 2015-2019 to capture a fuller picture of the demographic patterns in the city. Figure~\ref{fig:demographic} illustrates the spatial distributions of the two variables in 2019. There is apparent spatial clustering concerning how the marginalized population resides in the city. The original census data is at the census tract level, and to aggregate it to the taxi zone level for future estimation, we calculated the centroid of each census tract, assigned them to the corresponding taxi zone, and summed up the population of each demographic variable in each taxi zone. To calculate the ratio of each demographic, we divided the population of the target demographic population by the total population for each taxi zone.

\subsection{Performance Evaluation}
\subsubsection{Upstream Prediction Evaluation}

The study adopted two sets of metrics to evaluate the accuracy and fairness of the models respectively. To measure accuracy, we measured the error magnitude, relative error percentage, and error direction. Two commonly used metrics, Mean Absolute Error (MAE) and Root Mean Squared Error (RMSE), are adopted to evaluate error magnitude. They are defined as follows:
\begin{linenomath}
\begin{equation}
    MAE = \frac{1}{\kappa \times n} \sum_{k=1}^{\kappa} \sum_{i=1}^{n} |r_{i}^k - \hat{r}_{i}^k|,
\end{equation}
\end{linenomath}
\begin{linenomath}
\begin{equation}
    RMSE = \sqrt{\frac{1}{\kappa \times n} \sum_{k=1}^{\kappa} \sum_{i=1}^{n} (r_{i}^k - \hat{r}_{i}^k) ^2}.
\end{equation}
\end{linenomath}
In addition, to measure the relative error percentage, we adopted the Mean Absolute Percentage Error (MAPE):
\begin{linenomath}
\begin{equation}
    MAPE = \frac{1}{\kappa \times n} \sum_{k=1}^{\kappa} \sum_{i=1}^{n} \frac{|r_{i}^k - \hat{r}_{i}^k|}{r_{i}^{k'}} \text{, where }  r_{i}^{k'} = min(r_{i}^k, 0.1).
\end{equation}
\end{linenomath}
Note that the denominator $r_{i}^{k'}$ is adjusted to ensure the fraction is defined. Lastly, to measure the overall prediction error direction compared to the original demand, we adopted the Mean Error (ME):
\begin{linenomath}
\begin{equation}
    ME = \frac{1}{\kappa \times n} \sum_{k=1}^{\kappa} \sum_{i=1}^{n} (r_{i}^k - \hat{r}_{i}^k),
\end{equation}
\end{linenomath}
where a positive ME value denotes the model underestimating the demand in general, and vice versa.

On the other hand, to evaluate the fairness of the model predictions, we utilized two metrics. The first metric is the Mean-Variance of the Percentage Error (MVPE). This metric originates from the goal to ensure equal error distribution for all predictions. Given the percentage error in vertex $i$ at time $t$, $PE_{i}^k = \frac{r_{i}^k - \hat{r}_{i}^k}{r_{i}^k}$, the MVPE is defined as:
\begin{linenomath}
\begin{equation}
    MVPE = \frac{1}{\kappa} \sum_{k=1}^{\kappa} \frac{\sum_{i=1}^n (PE_{i}^k - \Bar{PE^k})^2}{n-1}
\end{equation}
\end{linenomath}
where $\Bar{PE^k}$ is the mean of the percentage errors in a given period $k$.

The second fairness metric refers to the Generalized Entropy Index proposed by \citep{speicher_unified_2018}. It is a metric originated from economics that measures inequality, or in the point of information theory, it can be interpreted as the $redundancy$ in data. In our case, it measures the spread between individual prediction and the average prediction error. A smaller value represents a more even distribution between errors and their mean. We adjusted the metric to evaluate the distribution of the percentage errors in our research setting:
\begin{linenomath}
\begin{align}
    & b_{i}^k = PE_{i}^k + m \text{, where } m \in \mathbb{R}^+ s.t. \forall PE_{i}^k, b_{i}^k \geq 0,  \\
    & GEI = \frac{1}{\kappa} \sum_{k=1}^{\kappa} [\frac{1}{n\alpha(\alpha-1)}\sum_{i=1}^n [(\frac{b_{i}^k}{\Bar{b^k}})^\alpha - 1]].
\end{align}
\end{linenomath}

Specifically, a temporal GEI for each period is first calculated and then we take the average of them across all periods. $\Bar{b^k}$ is the mean of $b_{i}^k$ in a given period $k$. $\alpha$ is a constant that regulates how much attention is put on the larger percentage of errors, and here we set $\alpha = 2$. 

\subsubsection{Downstream Vehicle Rebalancing Evaluation}

For the downstream performance evaluation of the vehicle rebalancing module, we adopt the ride-hailing simulator utilized in \citet{GUO2022}. 
For the simulation and evaluation of various approaches, demand data from June 26, 2019, is employed. The simulator is configured with a fleet size of 2000 vehicles. With this setup, vehicle and demand locations are initialized. At the start of the simulation, all vehicles are made available and evenly distributed across the taxi zones.

Since demand origins and destinations are at the sub-regional level, random road nodes within each sub-region are assigned as origins and destinations for customers in that particular sub-region. The simulator comprises two components: a matching engine, which is solved every 30 seconds, and a vehicle rebalancing engine, which is solved every 300 seconds. In the vehicle rebalancing model \ref{eq:MIVR}, the parameters are set as $\alpha=1$ and $\beta=10^2$. The vehicle rebalancing problem takes into account $\kappa=6$ time intervals ahead for optimization.

The simulation yields several key metrics for evaluating different rebalancing models, including:
\begin{enumerate}
    \item Average customer wait time
    \item Standard deviation of customer wait time across zones
    \item Customer unsatisfaction rate
    \item Average non-occupied VMT
    \item Average number of rebalancing trips
\end{enumerate}
These metrics are essential in assessing the performance and effectiveness of various fairness-enhancing rebalancing approaches. The variation in customer wait times, as measured by standard deviation, reflects the fairness of services provided to customers. Meanwhile, other metrics are indicative of the system's overall efficiency.

\subsection{Upstream Demand Prediction Results}
In this section, we discuss the effect of the socio-demographically enriched adjacency matrix on the performance of the upstream prediction of the ride-hailing demand.

\subsubsection{Performance Summary}

In this section, we demonstrate the demand prediction performance of the SA-STGCN model and different combinations of regularization terms compared to the Historical Average model and the pure STGCN model as the baseline models. Specifically, the Historical Average model is defined by using the historical average of the demand on the same date of the week and time interval of the day in the training set to predict the testing data.

Table~\ref{tab:model_cross} shows the prediction performance comparison among the two benchmark models, the SA-STGCN model, and models with regularization terms implemented in terms of accuracy and fairness. The first four metrics in the table (\textit{i.e.} MAE, RMSE, MAPE, and ME) demonstrate the accuracy performance of the models, while the last two metrics (\textit{i.e.} MVPE and  GEI) illustrate the fairness performance. The first section of the table~\ref{tab:model_cross} illustrates how SA-STGCN compare with the baseline models, and the second, third, and fourth sections of table~\ref{tab:model_cross} refer to the performance of the models with regularization terms penalizing overestimation, limiting error distribution, and the models with both regularization terms with different weight parameters.

\begin{table}[h!]
\small
\centering
    \begin{tabular}{l|cccc|cc}
    \toprule
    {} &   MAE &  RMSE &  MAPE &    ME &  MVPE &  GEI $(10^{-4})$ \\
    \midrule

    \textbf{Historical Average}    & 3.699 & 5.293 & 0.721 & -0.885 & 8.347 &    5.463 \\
    \textbf{STGCN}                 & 3.130 & 4.334 & 0.738 & -0.046 & 9.054 &    5.948 \\
    \textbf{SA-STGCN}              & 3.128 & 4.352 & 0.693 &  0.204 & 8.002 &    5.230 \\
    \midrule
    \textbf{SA-STGCN ($\boldsymbol{\gamma = 0.01}$)}  & 3.109 & 4.318 & 0.713 &  0.118 & 8.624 &    5.657 \\
    \textbf{SA-STGCN ($\boldsymbol{\gamma = 0.03}$)} & 3.113 & 4.350 & 0.659 &  0.463 & 7.426 &    4.846 \\
    \textbf{SA-STGCN ($\boldsymbol{\gamma = 0.05}$)} & 3.138 & 4.396 & 0.606 &  0.733 & 6.170 &    4.003 \\
    \textbf{SA-STGCN ($\boldsymbol{\gamma = 0.07}$)} & 3.173 & 4.458 & 0.577 &  0.988 & 5.637 &    3.648 \\
    \textbf{SA-STGCN ($\boldsymbol{\gamma = 0.09}$)} & 3.216 & 4.522 & 0.549 &  1.206 & 5.045 &    3.257 \\
    \midrule
    \textbf{SA-STGCN ($\boldsymbol{\lambda = 0.5}$)} & 3.152 & 4.360 & 0.744 & -0.057 & 9.004 &    5.917 \\
    \textbf{SA-STGCN ($\boldsymbol{\lambda = 1}$)} & 3.159 & 4.376 & 0.760 &  0.064 & 9.183 &    6.038 \\
    \textbf{SA-STGCN ($\boldsymbol{\lambda = 3}$)} & 3.197 & 4.415 & 0.728 &  0.108 & 8.143 &    5.320 \\
    \textbf{SA-STGCN ($\boldsymbol{\lambda = 5}$)} & 3.255 & 4.481 & 0.739 &  0.240 & 7.973 &    5.198 \\
    \midrule
    \textbf{SA-STGCN ($\boldsymbol{\lambda = 0.5, \gamma = 0.09}$)} & 3.136 & 4.359 & 0.723 &  0.143 & 8.934 &    5.863 \\
    \textbf{SA-STGCN ($\boldsymbol{\lambda = 1, \gamma = 0.07}$)} & 3.228 & 4.489 & 0.548 &  0.974 & 4.751 &    3.060 \\
    \textbf{SA-STGCN ($\boldsymbol{\lambda = 3, \gamma = 0.03}$)} & 3.346 & 4.575 & 0.568 &  0.786 & 4.540 &    2.929 \\
    \textbf{SA-STGCN ($\boldsymbol{\lambda = 5, \gamma = 0.07}$)} & 3.548 & 4.850 & 0.510 &  1.486 & 3.324 &    2.132 \\
    \bottomrule
    \end{tabular}
\caption{Model cross-comparison.}
\label{tab:model_cross}
\end{table}

As shown in the first section of the table~\ref{tab:model_cross}, it is evident that using the socio-demographically enriched matrix helps to improve both the accuracy and the fairness metrics compared the benchmarks. The SA-STGCN is able to reduce both the MVPE and GEI by 11.6\% and 12.1\% while not harming the accuracy metrics. It even slightly improved the MAE and the MAPE results. 

By cross-comparing the evaluation metrics of the performance of different regularization terms, the proposed regularization terms improve overall fairness significantly while not sacrificing model accuracy too much. In fact, the accuracy metrics in all three regularization models are improved in most cases compared to benchmarks. With extremely large weight parameters such as when $\lambda=5$ and $\gamma = 0.07$, the fairness metrics MVPE and GEI can be reduced by 63.3\% and 64.2\%, respectively, while the accuracy metrics such as MAE only increased by 13.4\% compared to the pure STGCN. Moreover, the positive outcomes of the relative error metric ME show that adding the regularization helps the prediction to shift from overestimation to underestimation. This transformation, as we will later illustrate, contributes significantly to the enhancement of vehicle re-balancing downstream. The model with overestimation estimation performs the best as it improves both fairness and accuracy metrics at the same time.

The second and third sections of the table \ref{tab:model_cross} also serve as the sensitivity analysis of the regularization weight's effect on the model performances. In the case of overestimation penalization, there is a consistent accuracy-fairness trade-off pattern, as the weight of regularization increases, the error magnitudes monotonically increase, the percentage error decreases, and the fairness improves. On the other hand, when restricting the error distribution, as we increase the magnitude of the weight of the regularization term, only the MAE, RMSE, and ME monotonically increase, but the MAPE and the two fairness metrics first increase then decrease. 

\subsubsection{Fairer Prediction in Space}

\begin{figure}[!h]
\centering
\begin{subfigure}{\textwidth}
  \centering
  \includegraphics[width=.9\linewidth]{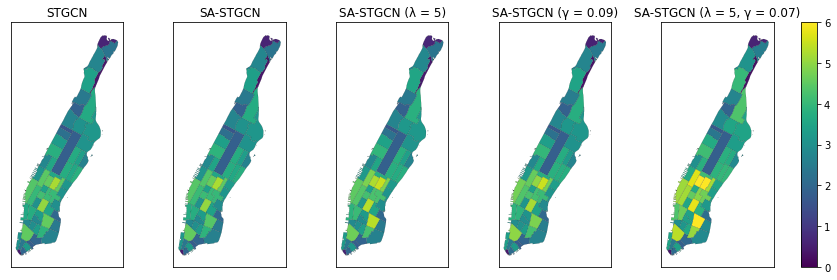}  
  \caption{MAE spatial distribution}
  \label{fig:mae_spatial}
\end{subfigure}
\begin{subfigure}{\textwidth}
  \centering
  \includegraphics[width=.9\linewidth]{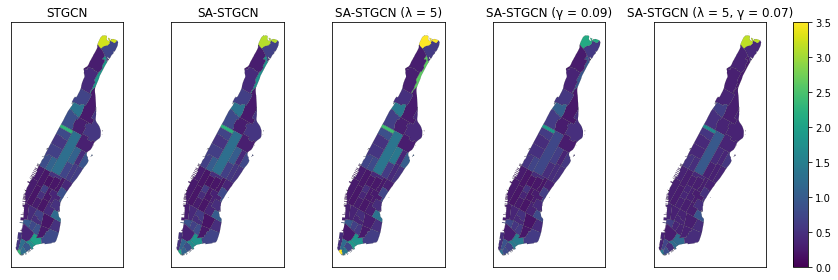}  
  \caption{MAPE spatial distribution}
  \label{fig:mape_spatial}
\end{subfigure}
\begin{subfigure}{\textwidth}
  \centering
  \includegraphics[width=.9\linewidth]{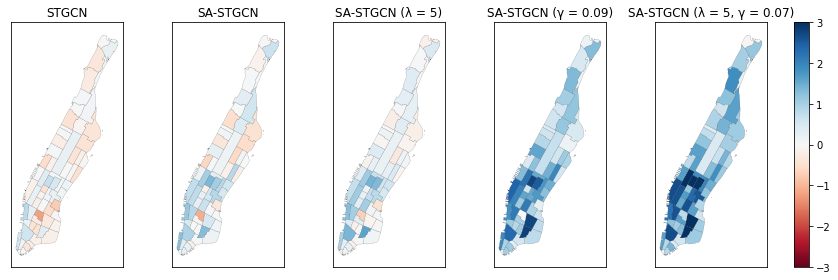}  
  \caption{ME spatial distribution}
  \label{fig:me_spatial}
\end{subfigure}
\caption{Error spatial distribution in Manhattan.}
\label{fig:error_spatial}
\end{figure}

Figure \ref{fig:error_spatial} demonstrates the spatial distribution of the error metrics across the study area of the pure STGCN, SA-STGCN, and the selected models with regularization terms. In general, the prediction errors are larger in magnitude in the downtown regions, while the percentage errors have more variation in the regions with low demand, such as the northern tip of the island. When comparing various models, the MAE plots reveal similar distributions, indicating that the prediction performances are not highly sensitive to the proposed changes in the model. In contrast, the MAPE plots demonstrate that models incorporating regularization terms exhibit a more uniform error distribution across space, reflecting a fairer prediction. The ME distributions suggested the general error direction shifted from negative to positive as regularization terms are introduced, which means a transition from overestimation to underestimation. We later demonstrate that this shift could be beneficial for promoting fairness in downstream vehicle rebalancing operations.

\subsubsection{Fairer Prediction in Time}



\begin{figure}[!h]
\centering
\begin{subfigure}{.48\textwidth}
  \centering
  \includegraphics[width=.95\linewidth]{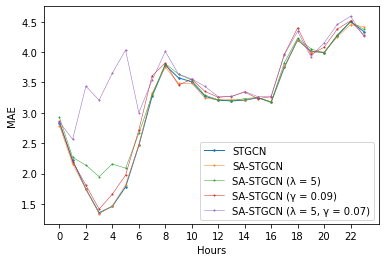}
  \caption{MAE temporal distribution}
  \label{fig:mae_temporal}
\end{subfigure}
\begin{subfigure}{.48\textwidth}
  \centering
  \includegraphics[width=.95\linewidth]{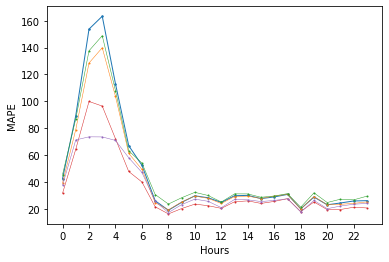}  
  \caption{MAPE temporal distribution}
  \label{fig:mape_temporal}
\end{subfigure}
\begin{subfigure}{.48\textwidth}
  \centering
  \includegraphics[width=.95\linewidth]{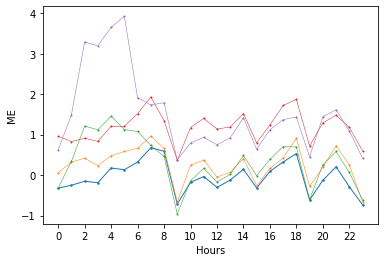}  
  \caption{ME temporal distribution}
  \label{fig:me_temporal}
\end{subfigure}
\begin{subfigure}{.48\textwidth}
  \centering
  \includegraphics[width=.95\linewidth]{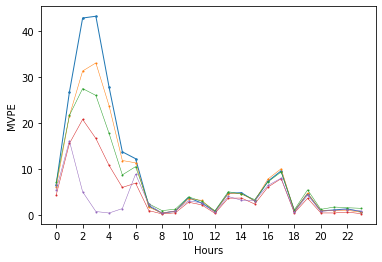}  
  \caption{MVPE temporal distribution}
  \label{fig:me_temporal}
\end{subfigure}
\caption{Error temporal distribution across times of the day.}
\label{fig:error_temporal}
\end{figure}

Figure \ref{fig:error_temporal} illustrates the temporal distribution of the error metrics across the time of the day. In general, the error magnitude and percentage distribution pattern follow the demand level, with smaller MAE values and larger MAPE and MVPE when demand is low and vice versa. In contrast, the error direction displays a less regular pattern. Compared to the benchmark performance of the pure STGCN model, the proposed SA-STGCN model and the method of adding regularization terms can reduce the peak errors in all scenarios. Specifically, SA-STGCN adding both regularization terms has the most direct impact on smoothing the MAE, MAPE, and MVPE temporal distribution. The smoothing effect is most significant in the early morning time when error values are more extreme. In addition, regularization terms have various effects on the error directions across time. When we add both regularization terms, the ME distribution shifts to positive the most as shown by the purple line. On the other hand, adding the error restriction regularization term has the least effect on day-time ME distribution. The temporal trend of RMSE is similar to MAE and the GEI trend is similar to that of MVPE, so the figures didn't include these two metrics.

\subsection{Downstream Vehicle Rebalancing Performances}

In this section, we discuss the effectiveness of our models when applied to vehicle reallocation tasks, utilizing a simulator based on real-world ride-hailing data.

\subsubsection{Performance Summary}

\begin{table}[hb!]
    \small
    \centering
    \resizebox{\textwidth}{!}{
    \begin{tabular}{l | ccccc}
        \toprule 
        & Unsatisfaction & Wait Time & Wait Time & Non-occupied & Rebalancing Trip \\
        & Rate (\%) & Avg (seconds) & Std (seconds) & VMT (miles) & Number \\
       \midrule 
       \textbf{True Demand} & 1.76 & 89.65 & 22.07 & 60.82 & 20.03 \\
       \textbf{Historical Demand} & 1.75 & 90.55 & 22.76 & 60.81 & 19.47 \\
       \midrule \midrule
       \textbf{Baseline STGCN} & 1.79 & 89.83 & 22.52 & 60.34 & 19.40 \\
       \textbf{Baseline SA-STGCN} & 1.74 & 89.75 & 22.13 & 60.20 & 19.25 \\
       \midrule \midrule
       \textbf{SA-STGCN ($\boldsymbol{\gamma = 0.01}$)} & 1.77 & 89.88 & 22.08 & 60.49 & 19.50 \\
       \textbf{SA-STGCN ($\boldsymbol{\gamma = 0.03}$)} & 1.76 & 89.95 & 22.53 & 60.29 & 19.25 \\
       \textbf{SA-STGCN ($\boldsymbol{\gamma = 0.05}$)} & 1.76 & 89.51 & 21.72 & 60.21 & 19.36 \\
       \textbf{SA-STGCN ($\boldsymbol{\gamma = 0.07}$)} & 1.72 & 89.67 & 21.77 & 60.00 & 19.08 \\
       \textbf{SA-STGCN ($\boldsymbol{\gamma = 0.09}$)} & 1.73 & 89.58 & 21.78 & 60.00 & 19.15 \\
       \midrule \midrule
       \textbf{SA-STGCN ($\boldsymbol{\lambda = 0.5}$)} & 1.73 & 90.16 & 22.34 & 60.37 & 19.28 \\
       \textbf{SA-STGCN ($\boldsymbol{\lambda = 1}$)} & 1.77 & 90.03 & 22.17 & 60.34 & 19.28 \\
       \textbf{SA-STGCN ($\boldsymbol{\lambda = 3}$)} & 1.78 & 89.98 & 22.15 & 60.36 & 19.33 \\
       \textbf{SA-STGCN ($\boldsymbol{\lambda = 5}$)} & 1.82 & 89.99 & 21.75 & 60.15 & 19.14 \\
       \midrule \midrule
       \textbf{SA-STGCN ($\boldsymbol{\lambda = 0.5, \gamma = 0.09}$)} & 1.75 & 89.39 & 21.06 & 59.87 & 19.09 \\
       \textbf{SA-STGCN ($\boldsymbol{\lambda = 1, \gamma = 0.07}$)} & 1.77 & 89.70 & 21.72 & 60.04 & 19.17 \\
       \textbf{SA-STGCN ($\boldsymbol{\lambda = 3, \gamma = 0.03}$)} & 1.73 & 90.02 & 22.10 & 59.99 & 18.90 \\
       \textbf{SA-STGCN ($\boldsymbol{\lambda = 5, \gamma = 0.07}$)} & 1.74 & 90.44 & 20.62 & 60.11 & 18.70 \\
        \bottomrule
    \end{tabular}}
    \caption{Model performance summary.}
    \label{tab:model_results_summary_with_weights}
\end{table}

Table \ref{tab:model_results_summary_with_weights} presents a summary of how various demand models perform in addressing the vehicle rebalancing problem, specifically when employing the fairness-focused MIVR model. This evaluation includes scenarios where true demand is known and those relying solely on average historical demand for future predictions, serving as benchmark models. Additionally, a standard STGCN and a baseline SA-STGCN model without additional regularization terms are also employed for benchmarking purposes.

Moreover, the evaluation also encompasses three variations of the SA-STGCN model: (i) SA-STGCN with a focus on penalizing overestimation, (ii) SA-STGCN with limited error distribution, and (iii) SA-STGCN incorporating both regularizations. Crucial customer service metrics assessed include the rate of customer dissatisfaction, average waiting times, and the variability of these waiting times across different sub-regions. To demonstrate the effectiveness of the vehicle rebalancing algorithms in enhancing operational efficiency, we examine metrics like the average Vehicle Miles Traveled without passengers per vehicle and the average number of rebalancing trips each vehicle undertakes.

\subsubsection{Fair Prediction Leads to Fair Service}

The first significant insight from the simulation results is the superior performance of predictions that incorporate fairness considerations, compared to benchmark prediction models. These enhanced models excel in two main aspects: reducing average customer wait times and achieving a more consistent wait time across different sub-regions. This improvement suggests that factoring in fairness can lead to more efficient and equitable service outcomes.

Further analysis reveals that the SA-STGCN model, when compared to the baseline STGCN, is more effective in satisfying a larger number of customers while simultaneously reducing the average wait time. Notably, this model also succeeds in minimizing the variance in service provision across different regions, addressing the issue of regional disparities in service quality.

The benefits of the SA-STGCN model stem from two primary factors. Firstly, it outperforms the traditional STGCN model in terms of both prediction accuracy and fairness indicators. These enhanced performances translate into more effective rebalancing operations, ensuring a fairer distribution of services with less disparity in error rates across regions. Secondly, the integration of socio-demographic information within the STGCN framework leads to a more conservative approach in demand estimation. This conservatism, while acknowledging the inherent inaccuracies in demand prediction, results in fewer non-occupied VMT and a reduced number of rebalancing trips. Consequently, this conservative approach benefits the overall system operation.

Moreover, the incorporation of additional regularization terms in the SA-STGCN models could potentially enhance model performance further. This improvement would be due to the reasons discussed above, emphasizing that fairer predictions not only improve service efficiency but also contribute to equitable service distribution. Ultimately, the proposed fairness-enhancing vehicle rebalancing approach fosters a win-win scenario, enhancing both the efficiency and fairness of ride-hailing services.

\subsubsection{The Power of Demand Underestimation}

The second major insight from the simulation results highlights the significant role of demand underestimation in ride-hailing rebalancing operations. This underestimation, prompted by the addition of a penalization term (associated with the parameter \(\gamma\)) to errors in demand overestimation, results in more equitable predictions. Despite a decrease in prediction accuracy, this approach leads to improved service provision for customers and reduces disparities in services across regions.

The advantage of demand underestimation lies in its inherent conservativeness. As \citet{GUO2022} suggest, in scenarios where future demand is uncertain, adopting a less aggressive approach often yields better outcomes. This conservatism allows for a more efficient distribution of limited vehicle resources, particularly important given the high costs associated with incorrect rebalancing decisions.

However, excessively penalizing demand overestimation in ride-hailing operations is not ideal. The simulation indicates that the optimal balance is achieved when \(\gamma\) is set to 0.05. This parameter setting results in the best average customer wait times and the fairest service distribution across different regions. It's important to note that while increasing \(\gamma\) does lead to fairer predictions, the impact on downstream services is not monotonic. Excessive conservatism in vehicle rebalancing can be counterproductive. Therefore, identifying an appropriate level of conservativeness is crucial for making optimal rebalancing decisions.

\subsubsection{The Most Fair Model}

In the comparative analysis of models presented in Table \ref{tab:model_results_summary_with_weights}, the SA-STGCN model with \(\lambda = 5\) and \(\gamma = 0.07\) emerges as the most equitable in terms of service provision to customers. This model significantly reduces the variability in customer wait times by 8.43\% compared to the baseline STGCN model, albeit at a slight increase in average wait times of 0.68\%. Notably, this model also excels in fairness regarding demand prediction, achieving the lowest MVPE of 3.32 and GEI of 2.13. These results underscore the correlation between fairer prediction models and more equitable service provision in the downstream.

Conversely, when considering a balance between efficiency and fairness in vehicle rebalancing, the SA-STGCN model with \(\lambda = 0.5\) and \(\gamma = 0.09\) stands out. In comparison to the baseline STGCN model, it successfully lowers both the standard deviation and average of customer wait times by 6.48\% and 0.49\%, respectively. Impressively, this model even outperforms scenarios with perfect demand knowledge in terms of average customer wait times. This can be attributed to the fairness-enhancing MIVR model, which approximates the matching component to account for future time intervals. Such approximations, while not perfectly optimal for real-world ride-hailing scenarios, demonstrate that sometimes, less accurate demand predictions can paradoxically lead to more efficient system operation.

\subsubsection{Trade-offs of Introducing Fairness Weights}

\begin{table}[h!]
    \small
    \centering
    \resizebox{\textwidth}{!}{
    \begin{tabular}{l | ccccc }
        \toprule 
        & Unsatisfaction & Wait Time & Wait Time & Non-occupied & Rebalancing Trip \\
        & Rate (\%) & Avg (seconds) & Std (seconds) & VMT (miles) & Number \\
       \midrule 
       \textbf{True Demand} & 1.81 & 89.49 & 22.55 & 60.70 & 20.13 \\
       \textbf{Historical Demand} & 1.78 & 90.51 & 22.62 & 60.73 & 19.53 \\
       \midrule \midrule
       \textbf{Baseline STGCN} & 1.79 & 89.79 & 22.80 & 60.36 & 19.50 \\
       \textbf{Baseline SA-STGCN} & 1.75 & 89.74 & 22.69 & 60.15 & 19.23 \\
       \midrule \midrule
       \textbf{SA-STGCN ($\boldsymbol{\gamma = 0.01}$)} & 1.79 & 89.87 & 22.78 & 60.29 & 19.38 \\
       \textbf{SA-STGCN ($\boldsymbol{\gamma = 0.03}$)} & 1.77 & 90.11 & 22.58 & 60.29 & 19.21 \\
       \textbf{SA-STGCN ($\boldsymbol{\gamma = 0.05}$)} & 1.75 & 89.57 & 22.28 & 60.04 & 19.25 \\
       \textbf{SA-STGCN ($\boldsymbol{\gamma = 0.07}$)} & 1.73 & 89.50 & 22.21 & 59.82 & 19.02 \\
       \textbf{SA-STGCN ($\boldsymbol{\gamma = 0.09}$)} & 1.76 & 89.32 & 22.56 & 59.88 & 19.16 \\
       \midrule \midrule
       \textbf{SA-STGCN ($\boldsymbol{\lambda = 0.5}$)} & 1.79 & 89.77 & 22.37 & 60.17 & 19.30 \\
       \textbf{SA-STGCN ($\boldsymbol{\lambda = 1}$)} & 1.76 & 90.14 & 22.59 & 60.21 & 19.13 \\
       \textbf{SA-STGCN ($\boldsymbol{\lambda = 3}$)} & 1.78 & 90.02 & 23.01 & 60.07 & 19.10 \\
       \textbf{SA-STGCN ($\boldsymbol{\lambda = 5}$)} & 1.77 & 89.80 & 22.50 & 60.03 & 19.14 \\
       \midrule \midrule
       \textbf{SA-STGCN ($\boldsymbol{\lambda = 0.5, \gamma = 0.09}$)} & 1.74 & 89.21 & 21.84 & 59.89 & 19.24 \\
       \textbf{SA-STGCN ($\boldsymbol{\lambda = 1, \gamma = 0.07}$)} & 1.80 & 89.70 & 21.96 & 59.83 & 19.05 \\
       \textbf{SA-STGCN ($\boldsymbol{\lambda = 3, \gamma = 0.03}$)} & 1.78 & 90.03 & 22.35 & 60.03 & 18.99 \\
       \textbf{SA-STGCN ($\boldsymbol{\lambda = 5, \gamma = 0.07}$)} & 1.75 & 90.35 & 21.79 & 59.91 & 18.63 \\
        \bottomrule
    \end{tabular}}
    \caption{Model performance summary without considering fairness weights in the MIVR model.}
    \label{tab:model_results_summary_without_weights}
\end{table}

Finally, we explore the implications of incorporating fairness weights into the MIVR model. Table \ref{tab:model_results_summary_without_weights} presents the performance metrics of the MIVR model sans the application of fairness weights in determining vehicle rebalancing strategies.

When analyzing the results from both tables, two significant insights are drawn. First, incorporating fairness weights into the MIVR model notably improves fairness, especially by diminishing discrepancies in customer wait times. Additionally, this approach results in serving a higher proportion of customers. This improvement is attributed to better servicing of customers in previously underserved areas, as more vehicles are strategically repositioned there. Evidence of this is seen in the increased average distance traveled by non-occupied vehicles (VMT) and the reduced number of rebalancing trips, which supports our hypothesis. However, the second observation reveals a compromise: while fairness is enhanced, there is an observable rise in the average customer wait time across the system.

\section{Policy Discussion} \label{sec:policy_discussion}
On October 30, 2023, President Joe Biden issued an executive order focusing on ``safe, secure, and trustworthy artificial intelligence'' \cite{house2023fact}, emphasizing the critical need to address the risks associated with artificial intelligence and establish new standards for its safety and security. This study reveals fairness issues within ride-hailing service algorithms, prompting consideration for regulatory interventions that would compel ride-hailing companies to integrate fairness considerations into their algorithmic design. For instance, governments can mandate that the variation in wait times across neighborhoods should not exceed a predefined threshold. Governments may also prescribe specific weights $\omega_i$ for individual regions within the vehicle rebalancing objective function (Equation \ref{eq:MIVR}), thereby exerting control over the importance assigned to successful matching in each distinct region $i$.

Concerning ride-hailing companies, the fairness-enhancing algorithm introduced in this research offers an effective solution to improve fairness in ride-hailing service provision. By adopting the proposed fairness-enhancing strategies in both demand prediction and service rebalancing, ride-hailing companies can greatly reduce variations in passenger wait times across the region. It’s worth highlighting that with the proper selection of hyperparameters $\lambda$ and $\gamma$ in the objective function of demand prediction (Equation \ref{eq:demand_obj}), both the variation and average customer wait time can be reduced. This achievement positions ride-hailing service companies to attain a Pareto improvement, concurrently enhancing both efficiency and fairness in their operations and achieving a ``win-win'' scenario.

To enhance equity in vehicle distribution, ride-hailing services can adopt strategies like Lyft's 'Power Zones.' These zones, located in underserved areas, offer drivers bonuses for accepting rides originating from these regions. This approach addresses the income disparity caused by low-demand zones, where surge pricing is less frequent and earning opportunities are diminished. By incentivizing drivers to operate in these areas, companies can not only ensure better service coverage but also support drivers in maintaining consistent earnings.


From the perspective of passengers, this approach, which aims for a more equitable distribution of wait times among communities, especially benefits individuals facing challenges with transportation options. By fine-tuning both demand and rebalancing models, it has the potential to simultaneously attain a lower and more equitable wait time distribution among people from different communities. This improvement enhances mobility and connectivity across all communities, fostering trust among passengers in the ride-hailing service, especially within disadvantaged communities.

The ride-hailing service plays a pivotal role as a transportation solution for individuals seeking employment and opportunities. The decrease in wait times translates to more efficient access to opportunities for residents in diverse communities, offering significant relevance for areas with elevated unemployment rates or limited local job options. Through the adoption of this fairness-enhancing algorithm, ride-hailing companies not only showcase their commitment to social responsibility but also actively contribute to the overall well-being and inclusivity of the communities they serve.

\section{Conclusion} \label{sec:summary}

This study presents a pioneering framework aimed at enhancing fairness in both predicting ride-hailing demand and delivering equitable service to riders. The framework introduces a Socio-Aware Spatio-Temporal Graph Convolutional Network (SA-STGCN), which integrates a socio-enriched adjacency matrix and bias-reduction regularization methods. Additionally, it features a vehicle rebalancing engine that incorporates fairness considerations into its objective function. This framework was evaluated using a simulator with real-world ride-hailing data, demonstrating that the SA-STGCN model not only outperforms standard demand prediction models in accuracy but also in fairness. Significantly, improvements in fairness at the demand prediction stage lead to more equitable service delivery in the vehicle rebalancing process. The vehicle rebalancing module, enhanced with fairness weights, showed a notable reduction in both the standard deviation and average of customer wait times by 6.48\% and 0.49\%, respectively, suggesting a win-win scenario where both efficiency and fairness are improved for ride-hailing platforms.

The proposed framework offers a viable approach for ride-hailing companies to integrate fairness into their operations, and it provides a basis for government regulations aimed at preventing service imbalances across different areas. However, realizing the win-win scenario highlighted in the study involves addressing practical challenges. A key solution lies in developing driver incentive mechanisms. These mechanisms should ensure that drivers are motivated to serve in underserved communities and that their earnings remain stable despite such commitments. As the role of ride-hailing services becomes more central in our everyday activities, it's crucial to make certain that these platforms maintain a strong commitment to social responsibility and proactively enhance the well-being and inclusiveness of the communities they operate in.

For future research, it would be beneficial to include a focus on driver behaviors and earnings, which this study has not addressed. A comprehensive framework could be developed to enhance fairness across all aspects of the ride-hailing vehicle rebalancing operations: demand prediction, fairness for riders, and fairness for drivers. Such a framework should ensure that drivers who are redirected to serve underserved communities are compensated equitably, comparable to those serving in city centers. This approach would create a more balanced and fair environment for all parties involved in the ride-hailing ecosystem.

\section*{Acknowledgements} 

\bibliographystyle{model1-num-names}
\bibliography{reference.bib}

\newpage
\appendix

\end{document}